\documentclass[draft]{agujournal2019}
\pdfoutput=1
\usepackage{url} 
\usepackage{float}
\usepackage[inline]{trackchanges} 
\usepackage{soul}
\usepackage{amsmath}
\usepackage{subcaption}
\usepackage{booktabs}
\usepackage[colorinlistoftodos]{todonotes} 

\draftfalse

\journalname{JGR: Atmospheres}

\begin{document}

%
%


\title{CNCast: Leveraging 3D Swin Transformer and DiT for Enhanced Regional Weather Forecasting}

%
%




\authors{Hongli Liang\affil{1}, Yuanting Zhang\affil{1}\thanks{zhangyuanting@caiyunapp.com}, Qingye Meng\affil{1}, Shuangshuang He\affil{1}, Xingyuan Yuan\affil{1}}


\affiliation{1}{ColorfulClouds Technology Co.,Ltd}




\correspondingauthor{Yuanting Zhang}{zhangyuanting@caiyunapp.com}



\begin{keypoints}
\item Development of an advanced regional weather forecast model: This study introduces an advanced regional weather forecasting model leveraging the SwinTransformer 3D architecture. Designed to deliver precise hourly weather predictions beyond five days, this model enhances the reliability and usefulness of short-term meteorological forecasts, paticularly the wind forecasts.
\item Generally superior performance compared to a established global model: Results indicate that our model outperforms Pangu, a leading global weather forecasting system, across most weather variables. This demonstrates the model's potential to offer a more effective forecasting solution.
\item Integration of enhanced boundary conditions: Inspired by numerical weather prediction (NWP) methodologies, our model incorporates refined boundary conditions that significantly boost its predictive accuracy. 
\item  Diagnosis of hourly high resolution total precipitation: The model estimates hourly total precipitation at a high resolution of approximately 5 kilometers using a latent diffusion model. This provides an innovative method for generating high-resolution precipitation data, offering an alternative approach 
to traditional precipitation diagnosis techniques.
\end{keypoints}

%
%

%
%


\begin{abstract}

This paper presents an advanced regional weather forecasting model utilizing the SwinTransformer 3D architecture, which is designed to deliver accurate hourly predictions up to five-day horizon. This innovative model significantly enhances the reliability and accuracy of short-term meteorological forecasts. Through rigorous performance evaluations, our model has demonstrated superior capabilities compared to Pangu, the global weather forecasting model, and outperforming it across most variables. A key enhancement of this model is the integration of refined boundary conditions, inspired by traditional numerical weather prediction (NWP) methodologies. This strategic incorporation markedly improves predictive accuracy. Additionally, the model introduces an innovative approach for estimating hourly total precipitation at a high spatial resolution of approximately 5 kilometers using a latent diffusion model. This advancement provides a novel approach to generate high-resolution precipitation data. In general, this study highlights significant advancements in regional weather forecasting, offering enhanced precision and reliability to support various meteorological applications.

\end{abstract}

\section*{Plain Language Summary}
This study introduces a cutting-edge regional weather forecasting model based on the SwinTransformer 3D architecture. This model is specifically designed to deliver precise hourly weather predictions ranging from 1 hour to 5 days, significantly improving the reliability and practicality of short-term weather forecasts. Our model has demonstrated generally superior performance when compared to Pangu, a well-established global model. The evaluation indicates that our model excels in predicting most weather variables, highlighting its potential as a more effective alternative in the field of limited area modeling. A noteworthy feature of this model is the integration of enhanced boundary conditions, inspired by traditional numerical weather prediction (NWP) techniques. This integration has substantially improved the model's predictive accuracy. Additionally, the model includes an innovative approach for diagnosing hourly total precipitation at a high spatial resolution of approximately 5 kilometers. This is achieved through a latent diffusion model, offering an alternative method for generating high-resolution precipitation data.

%
%

%


%
%
%
%

\section{Introduction}

Accurate and reliable weather forecasting is crucial for a wide range of applications, including agriculture, disaster management, transportation and daily planning. Traditional forecasting methods often face challenges in providing precise short-term predictions, particularly at regional scales. Traditional numerical weather prediction(NWP) models construct differential equations for physical forecasting and solve them using numerical methods to obtain future weather forecasts\cite{SemiLagrangianMethod, ec_program_2020}. In addition, total precipitation is diagnosed through different types of parametarization schemes. Due to the complexity of these equations and the numerous parameters involved, numerical models require substantial supercomputing resources and tend to have slower solution speeds\cite{nwp_evolution, nwp_evolution_2019}. As a result, the release of real-time forecast data is often delayed by several hours, and the spin-up process also leads to relatively poorer performance in the first few hours of the forecast. Furthermore, for weather predictions at sub-grid scales, numerical models rely on various parameterization schemes\cite{nwp_evolution}, which can introduce different types of errors when simulating small-scale weather phenomena.

In recent years, the rapid advancements in deep learning and high-performance GPUs and TPUs have initiated a new revolution in weather forecasting. Following the release of FourCastNet\cite{fourcastnet} for global forecasting, 2023 saw the introduction of PanguWeather, which became the first large-scale global weather forecasting model to surpass the ECMWF's IFS in some metrics\cite{pangu}. This breakthrough sparked a wave of research and the release of several large-scale weather models, including Graphcast, Fengwu, Fuxi, AIFS, ClimaX, and GenCast. Among these, GraphCast, Gencast and AIFS build weather forecasting models based on graph neural networks\cite{graphcast, gencast, aifs2024lang}, providing forecasts for several days ahead at six-hour intervals. Meanwhile, the FengWu, Fuxi, and ClimaX\cite{FengWuPT, fuxi2023, climax} series models are constructed using vision transformer-like architectures\cite{vit}. Training these models on global data demands significant computational resources, and some models exclude a crucial variable—precipitation. 

In the context of global weather forecasting model research trends, there has been relatively less focus on regional weather forecasting models. \citeA{metmamba} addresses this gap by constructing a limited area model based on the Mamba\cite{mamba, vmambaliu2024}, commonly found in large language models, providing 6-hourly weather forecasts for the next five days. Notably, this model does not include precipitation. To enable hourly precipitation forecasting, we follow the NWP approach by treating precipitation as a diagnostic variable, generated separately using the DiT model. DiT is a diffusion model with a vision transformer as its backbone\cite{dit}, which can effectively capture spatial dependencies and complex patterns in precipitation data. Diffusion models, which learn the distribution transformation from noise to data through a denoising reverse process\cite{diffusion_process, score_based_model_2019, song2021scorebased}, have become mainstream architectures in fields such as image and video generation\cite{stable_diffusion_2021, stable_video_diffusion}. There have been numerous studies on precipitation forecasting using diffusion models\cite{prediff_gao2023, precip_nowcast_latent_diffusion_2023, DGDM2023, precip_nowcast_diffusion_2024}, which have demonstrated that these models can enhance the accuracy and texture clarity of precipitation forecasts.

For variables such as temperature, pressure, humidity, and wind, this paper follows the Pangu approach, utilizing a modified Swin Transformer 3D model to achieve weather forecasts for these variables in the region of China. For the precipitation, we achieved high-resolution precipitation diagnosis using the latent DiT model. Ultimately, we developed a comprehensive regional weather forecasting system in China (CNCast) that outperforms Pangu in terms of RMSE and accuracy across most variables.

\section{Data Preparation}
\subsection{Regional ERA5}

For training and evaluating CNCast, we downloaded and cropped the global ERA5 dataset from the open Google Cloud bucket. ERA5 is the fifth generation of the ECMWF reanalysis dataset, produced using the ECMWF Integrated Forecast System (IFS) cycle 42r1, which was operational for most of 2016\cite{era5-total,era5-upper}. This dataset assimilates high-quality global observations, making it an excellent benchmark for global weather forecasting.

CNCast is trained on a subset of the highest spatial resolution available in the ERA5 dataset, using a $0.25^\circ$ equiangular grid that spans from $0^\circ$ to $60^\circ$ N latitude and from $70^\circ$ to $140^\circ$ E longitude. This results in an input resolution of $241\times281$ (241 for latitude and 281 for longitude). Following the approach in \cite{pangu}, we selected 4 surface variables and 5 pressure level variables at 13 pressure levels (100, 150, 200, 250, 300, 400, 450, 500, 600, 700, 850, 950 and 1000 hPa). The pressure level atmospheric variables are geopotential (z), temperature (t), u component of wind (u), v component of wind (v), and specific humidity (q). The surface variables include 2-meter temperature (2mt), 10-meter u wind component (u10), 10-meter v wind component (v10), and mean sea level pressure (mslp). Additionally, topography is included in the input data for surface variables.

From the 43 years of data, the years 1979 to 2019 are used for training, 2020 for validation, and 2021 for testing.

\subsection{Precipitation} 

Precipitation is a critical meteorological element with significant impacts on human life and societal activities. In addition to forecasting elements such as temperature, pressure, humidity, and wind, this paper also diagnoses precipitation based on these factors. In China, a high-resolution precipitation analysis product known as China Meteorological Precipitation Analysis System (CMPAS) has been developed by the National Meteorological Information Center(NMIC) of the China Meteorological Administration(CMA). This product integrates various precipitation data sources employing merging methods of 'PDF(Probability Density Function)+OI(Optimal Interpolation)'\cite{morph}. Combining the 'PDF+OI+BMA(Average Bayesian Model)+DS(Downscaling)' method, NMIC developed CMPAS\cite{cmpas1, cmpas2}, a precipitation estimation product fused with three sources (gauge, satellite, radar) at 1 km resolution. 

In this study, we leveraged the high-resolution CMPAS precipitation dataset for the task of precipitation diagnosis. The dataset offers hourly precipitation measurements with an impressive spatial resolution of $0.01^\circ \times 0.01^\circ$, covering latitudes from $15^\circ$N to $60^\circ$N and longitudes from $70^\circ$E to $140^\circ$E. This results in a grid comprising 4,500 latitude points and 7,000 longitude points.

To make the data more manageable and reduce computational costs, we applied a downsampling process to achieve a lower resolution of $0.05^\circ \times 0.05^\circ$. This was accomplished using a sliding average pooling technique, which allows us to effectively capture the essential features of the precipitation data while significantly reducing the size and complexity of the dataset. This approach ensures that our analysis remains computationally feasible without sacrificing the integrity of the precipitation information crucial for accurate diagnosis.

\subsection{Data Preprocessing}
For variables other than total precipitation, we applied z-score normalization using the mean and standard deviation calculated from the training dataset. For total precipitation, we took a different approach to address the challenges posed by its long-tailed distribution, which can hinder the model's ability to accurately forecast heavy precipitation events. To tackle this, we first transformed the precipitation values into dBZ (decibels relative to Z), a logarithmic scale commonly used in meteorological radar reflectivity data to better represent the intensity of precipitation. This transformation reduces the long tail and brings the data closer to a Gaussian distribution, making it more conducive to the training of DiT model.

After converting the precipitation data to dBZ, we then applied z-score normalization using the mean and standard deviation of the dBZ values derived from the training dataset. This two-step transformation allows the model to more effectively learn patterns associated with both typical and extreme precipitation events, ultimately enhancing its performance in forecasting heavy precipitation.

\section{Methodology}
\subsection{Problem Settings}

Regional weather forecasting systems that utilize deep learning frameworks are predominantly developed using analysis or reanalysis data. In this study, we employ the ERA5 dataset alongside CMPAS to implement CNCast training. CNCast is structured into two main components.

The first component focuses on weather forecasting using the Swin Transformer 3D. Unlike global weather forecasting, regional weather forecasting requires boundary conditions to ensure the stability and accuracy of predictions. To enhance model accuracy over longer forecast durations, we extracted 4 outmost pixels as boundary at lead time $t$ from the ERA5 data specific to China region and fed it to the model. Based on the ERA5 data, the regional weather forecasting problem is defined by Equation \ref{eq:fcst_stage1}: given an initial time $t$, the model $F$ with trainable parameters $\theta$ utilizes historical weather data $X_t$, topography $d$ and boundary conditions at time $t+s$(noted as $B_{t+s}$), to predict future weather state at time $t+s$(noted as $X_{t+s}$), with $s$ as interval. The hourly weather forecasting model is capable of predicting 69 meteorological variables over a five-day period. To achieve efficient iterative forecasting, we initially trained a 1-hour forecast model. Building on this model, we then fine-tuned it to create forecast models with lead times of 3, 6, and 24 hours. To optimize the forecasting process, we followed \citeA{pangu} and employed a greedy algorithm to minimize the number of iterative steps required, thereby facilitating hourly iterative forecasts for the next five days.

\begin{equation} 
    X_{t+s} = F(X_t, B_{t+s}, d; \theta) 
    \label{eq:fcst_stage1}
\end{equation}

The second component is designed for precipitation diagnostic forecasting. Similar to NWP, the current moment's precipitation is diagnosed based on the weather states $X_t$, ERA5 total precipitation $p_t$ and CMPAS precipitation at last hour $P_{t-1}$, with CMPAS precipitation at time $t$ serving as the target. To more effectively capture the intensity and structure of the precipitation, we utilize the conditional latent diffusion model\cite{ldm, dit}, denoted as $H$. As illustrated in Equation \ref{eq:fcst_stage2}, the precipitation at time $t$(noted as $P_t$) is retrieved by $H$, which has learnable parameters $\omega$, using encoded $X_t$, $p_t$ and $P_{t-1}$ as inputs. To reduce the compute cost for training diffusion model, we trained VAEs to compress and encode the ERA5 69 variables, ERA5 total precipitation and CMPAS precipitation, respectively. Therefore, $X_t$ is encoded by $V_x$, $P_{t-1}$ by $V_{cmpas}$ and $p_t$ by $V_p$, before being fed to $H$.

\begin{equation} 
    P_t = H(V_x(X_t), V_{cmpas}(P_{t-1}), V_p(p_t); \omega) 
    \label{eq:fcst_stage2}
\end{equation}

\subsection{Model Details}
\subsubsection{Weather Forecasts with SwinTransformer3D}

\begin{figure}[H]
    \centering
    \includegraphics[width=0.9\linewidth]{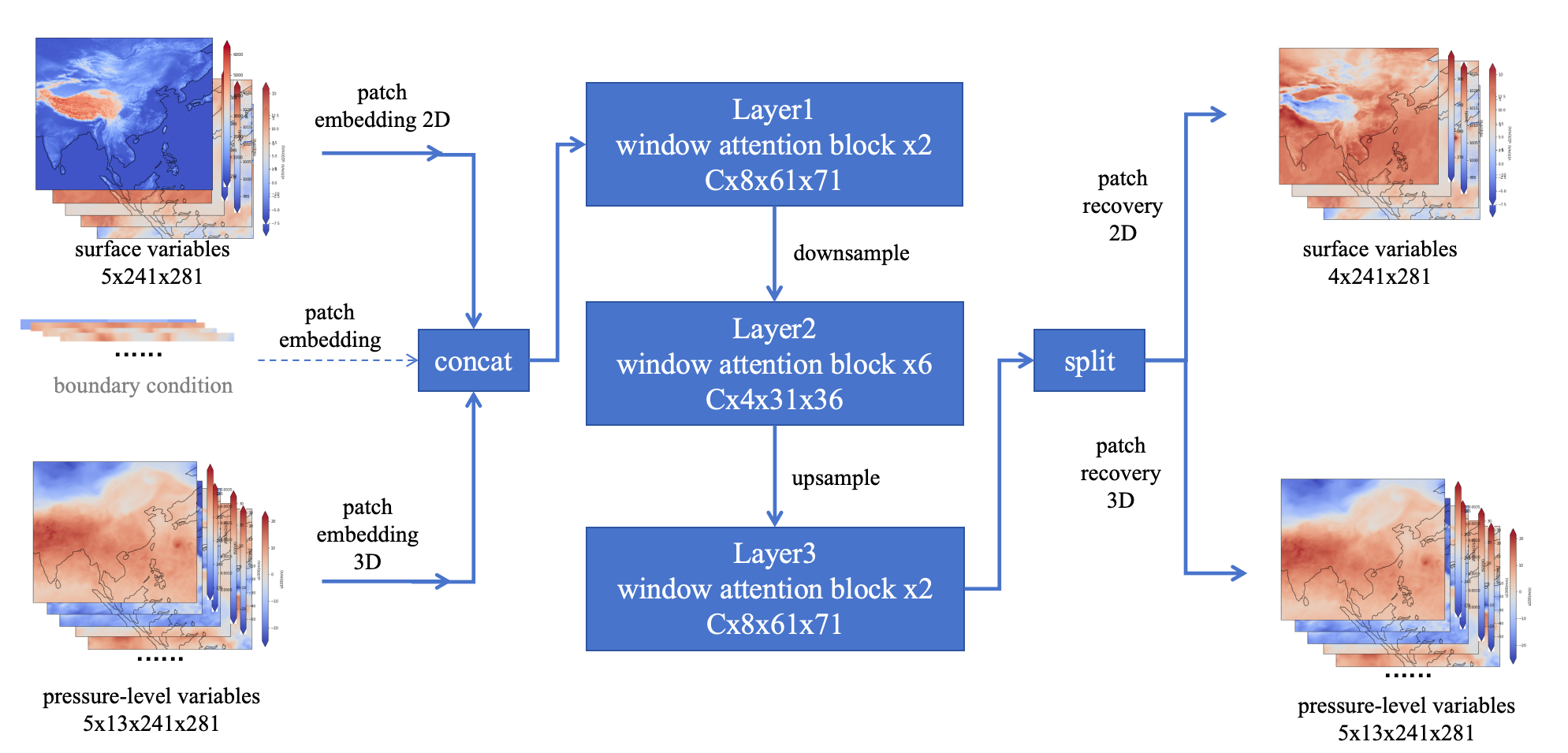}
    \caption{Swin Transformer 3D architecture.}
    \label{fig:swin3d}
\end{figure}

Based on the dataset described previously, we employed the Swin Transformer 3D model to forecast future weather states. The overall architecture of the model is depicted in Figure \ref{fig:swin3d}. At the initial time $t$, the model's input is composed of three components: surface variables, pressure level variables, and boundary conditions. Each of these components undergoes a separate patch embedding process with the following sizes: $4\times241\times281$ for surface variables, $65\times241\times281$ for pressure level variables, and $69\times4\times1044$ for boundary conditions. The model's output consists of 69 variables, each with a size of $241\times281$.

The model architecture comprises three layers. Hidden features are downsampled before being passed to the second layer and upsampled before being input into the third layer. By using such architecture, the model can capture and improve the critical features needed for accurate weather forecasts. 

\textbf{Sliding Patch Embedding}

Meteorological models that utilize Swin Transformer- and window attention-like architectures, such as Pangu and Fuxi, often follow the Vision Transformer (ViT) approach \cite{vit} for patch embedding, where the kernel size and patch size are typically identical. However, research has shown that using a larger kernel size can expand the receptive field\cite{large_kernel_cnn, large_kernel_matters}, and employing sliding patch embedding can reduce the checkerboard effect, thereby enhancing performance.

To take advantage of these findings, we increased the kernel size by 3 during the patch embedding process and implemented sliding patch embedding. Specifically, for patch embedding, the patch sizes for the surface and pressure level components were set to [4, 4] and [2, 4, 4], while the corresponding kernel sizes were set to [7, 7] and [5, 7, 7], respectively. 

We conducted an ablation study to verify the effectiveness of this approach. The results, presented in Table \ref{tab: sliding patch}, show that the latitude-weighted Root Mean Square Error (RMSE) for all variables is lower when using sliding patch embedding compared to not using it. This demonstrates the benefit of sliding patch embedding approach in improving model performance. Therefore, a larger kernel size is adopted as the default configuration in this paper.

\begin{table}
    \caption{Latitude weighted RMSE of surface and pressure level variables at 500hPa.}
    \centering
    \begin{tabular}{c c c c c c c c c c}
    \hline
      & t2m&mslp&u10&v10&t500&z500&q500&u500&v500 \\
    \hline
    w/o sliding patch & 1.110  & 51.781 & 0.423 & 0.420 &	0.242 &	36.555 &1.319e-4 &	0.588 &	0.543 \\
    w/ sliding patch& 1.024 &46.707 &0.376 &0.370 &	0.227 &33.728 &1.202e-4 &0.546 &0.504 \\
    \hline
    \end{tabular}
    \label{tab: sliding patch}
\end{table}

\textbf{Lateral Boundary Condition}

Unlike global weather forecasting, regional weather forecasting necessitates the use of lateral boundary conditions to constrain prediction results during extended forecasts, thereby preventing over-smoothing and divergence as lead time increases. To address this issue, \citeA{metmamba} utilized FourcastNet predictions as boundary to replace the outmost pixels of input features. While incorporating FourCast forecasts as boundary can enhance the model’s forecasting accuracy over long lead times, relying on predictions from a specific model for boundary widens the information disparity between the boundary and input data. This approach also hampers the model’s adaptability to alternative initial fields during operational implementation. Moreover, directly substituting boundary for the pixels surrounding the input leads to partial information loss, which adversely affects model training. In contrast to \citeA{metmamba}, our method extracts 4 outmost pixels surrounding China from ERA5 data as boundary and applies separate embedding to both the boundary and input before concatenation. This strategy preserves information integrity and effectively minimizes the information discrepancy between the boundary and input.

\begin{figure}[H]
    \centering
    \includegraphics[width=0.8\textwidth]{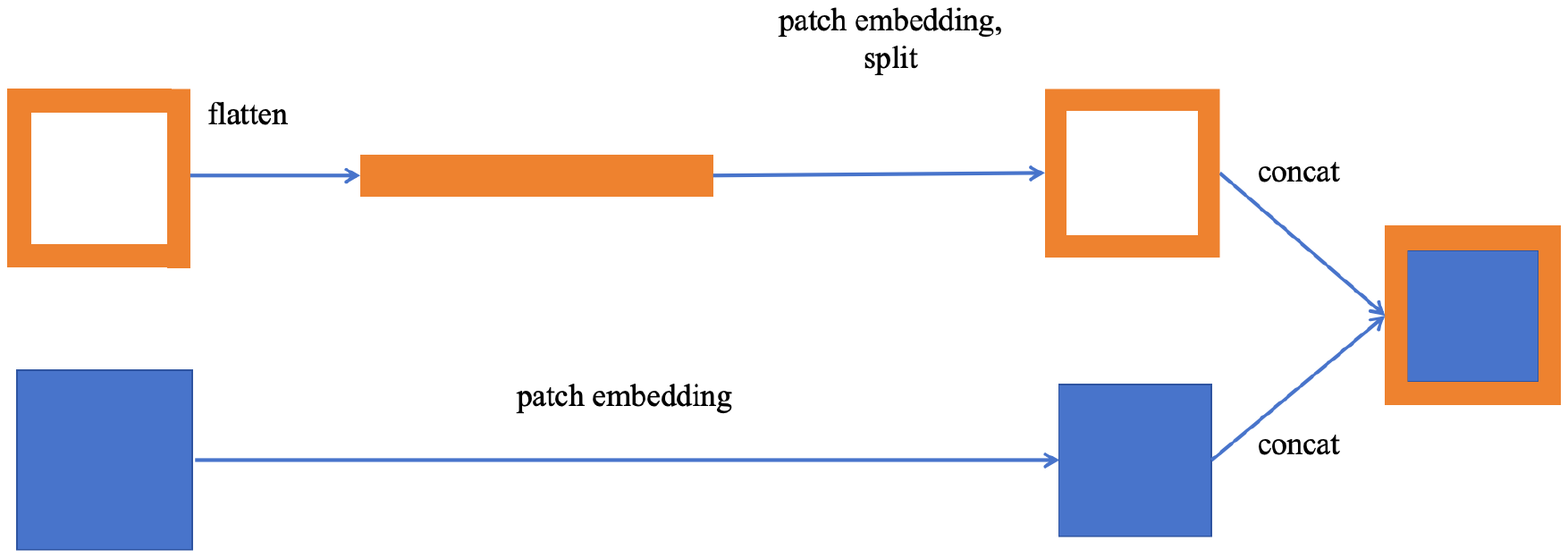} 
    \caption{Patch embedding with boundary condition. The boundary is taken from target with 4 outmost pixels and then flattened to be fed into patch embedding layer. The embedded boundary is then split to 4 parts and concatenated back to the embedded feature.}
    \label{fig:patch_embedding}
\end{figure}

As depicted in Figure \ref{fig:patch_embedding}, when feeding data into the Swin Transformer 3D model, we performed patch embedding on the boundary condition separately. The embedded boundary data was then concatenated with the embedded data of the original variables. This combined embedding was subsequently fed into the attention layers for feature extraction. Experimental results demonstrate that this method of incorporating boundary conditions effectively constrains forecast outcomes, ensuring stability and accuracy over longer lead times.

\subsubsection{Precipitation Diagnosis with DiT}  

In NWP, precipitation is typically diagnosed from other meteorological variables. Following this paradigm, we diagnose hourly precipitation based on variables such as temperature, mean sea level pressure, specific humidity, and wind. This approach allows us to forecast precipitation after predicting these variables using the Swin Transformer 3D model. 

To leverage diffusion model advantages, we employ a Diffusion-based Transformer (DiT) model for precipitation diagnosis. Given the high computational demands of diffusion models and ViT-based architectures, we first train three Variational Autoencoders (VAEs) to compress the ERA5 non-precipitation variables, as well as the ERA5 and CMPAS hourly accumulated precipitation. Subsequently, we train the DiT model using these latent representations. This approach allows us to manage computational resources efficiently while maintaining high prediction accuracy.

\begin{figure}[H]
    \centering
    \includegraphics[width=0.98\textwidth]{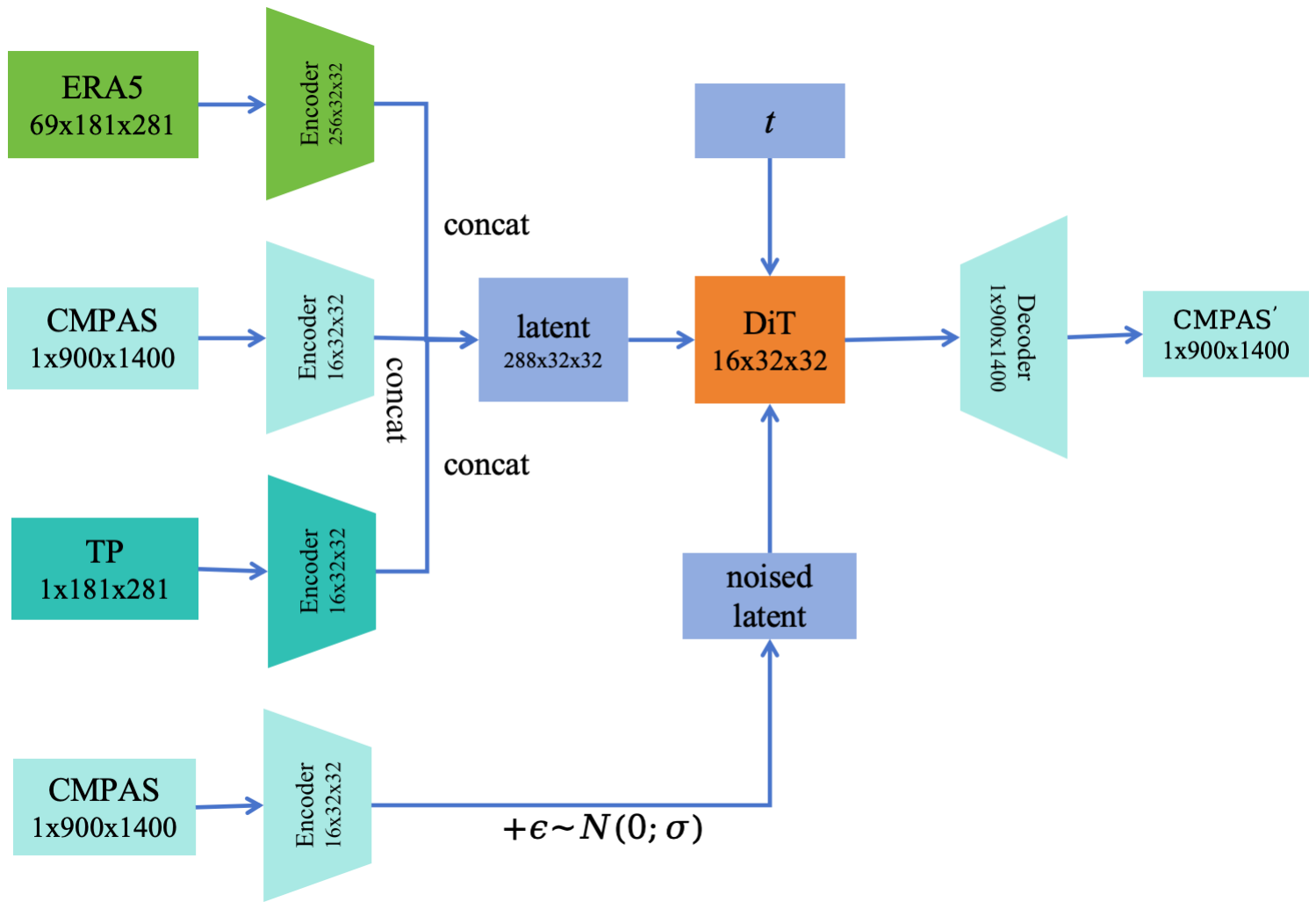} 
    \caption{Precipitation diagnosis with DiT based on latents.}
    \label{fig:precipitation_dit}
\end{figure}

The process of precipitation diagnosis is illustrated in Figure \ref{fig:precipitation_dit}. The surface and pressure level non-precipitation variables from ERA5, along with the ERA5 total precipitation, are compressed by Variational Autoencoders (VAEs) from dimensions of $69\times181\times281$ and $1\times181\times281$ to latent representations of $256\times32\times32$ and $16\times32\times32$, respectively. Similarly, the CMPAS data is compressed from $1\times900\times1400$ to $16\times32\times32$ using VAE.

Building on the latent representations generated by the VAEs, a conditional DiT-S-2 model \cite{dit} is trained to diagnose high-resolution precipitation. The model's input conditions include the latent representations of ERA5 non-precipitation variables, the ERA5 total precipitation at the current time step, the latent representation of CMPAS at the previous time step, as well as the noise schedule step $t_{noise}$. The target output is the latent representation of CMPAS at the current time step. The diffusion process follows the Denoising Diffusion Probabilistic Model (DDPM) \cite{ddpm} with a linear noise schedule. To follow \citeA{dit}, the input latent is decomposed into patches with a patch size of [2, 2] and an embedding dimension of 192, and then processed through 12 DiT blocks. During the training of DiT-S-2, the encoders and decoders of the VAEs remain frozen. High-resolution precipitation is achieved using the CMPAS decoder, allowing for precise and detailed precipitation diagnosis.

\subsection{Model Training}

To enhance the performance of the Swin Transformer 3D model, the Mean Squared Error (MSE) loss function was selected. For the DiT-S-2 model, a combination of MSE and Kullback-Leibler (KL) divergence was employed to model the noise distribution across varying noise schedules. Although the latent diffusion model excels at capturing and learning data distributions, its effectiveness is partially contingent on the quality of the latent representations generated by the VAEs. To minimize information loss and maintain data distribution integrity during compression, the training process of the VAEs is augmented with a Generative Adversarial Network (GAN). The generator is trained using a loss function composed of four components, as shown in Equation \ref{eq:loss_vae_g}.
\begin{itemize}
\item{Reconstruction Loss: Measured using Mean Absolute Error (MAE), it ensures that the reconstructed output closely matches the original input.}
\item {Kullback-Leibler (KL) Divergence: With a scaling factor ($\gamma$), this term encourages the latent space distribution to approximate a standard normal distribution, aiding in regularization.}
\item{Learned Perceptual Image Patch Similarity (LPIPS): With a scaling factor ($\lambda$), this component captures perceptual differences between the original and reconstructed images, enhancing the perceptual quality of the output.}
\item{Discriminated Loss: With a learnable scale ($\psi$), it leverages a discriminator to ensure that the generated outputs are indistinguishable from real data, improving the generative quality of the VAEs.}
These components collectively guide the generator towards producing high-quality, realistic latent representations while maintaining the integrity of the original data.
\end{itemize}

\begin{equation}
    L_G=MAE+\lambda*LPIPS+\gamma*KL+\psi*D(G(X))
    \label{eq:loss_vae_g}
\end{equation}

We employed the AdamW optimizer and established an initial learning rate of 3e-4 for both the Swin Transformer 3D and DiT-S-2 models. To speed up training, we leveraged PyTorch Distributed Data Parallel (DDP). The Swin Transformer 3D was trained on a 64-core TPU v4 card for 100 epochs, while DiT-S-2 was trained on two 40GB Tesla A100 cards for 70,000 iterations.

\subsection{Evaluation Metrics}

Following previous NWP-based and AI-based methods, we quantitatively evaluate the forecast(exclude precipitation) using two commonly applied metrics: latitude-weighted Root Mean Squared Error (RMSE) and Anomaly Correlation Coefficient (ACC). The latitude weighted RMSE measures the average magnitude of errors between the predicted and observed values, with lower values indicating better performance. It is defined as shown in Equation \ref{eq:weighted-rmse}. The latitude weighted ACC evaluates how well the forecast captures spatial anomaly patterns compared to a reference climatology. It assesses the phase agreement between forecasted and observed anomalies, with higher values indicating better performance. The ACC is defined as shown in Equation \ref{eq:weighted-acc}.
\begin{equation}
    \text{RMSE}_{\text{weighted}} = \sqrt{ \frac{1}{N} \sum_{i=1}^{N} w_i \cdot (y_i - \hat{y}_i)^2 }
    \label{eq:weighted-rmse}
\end{equation}
\begin{equation}
    \text{ACC}_{\text{weighted}} = \frac{ \sum_{i=1}^{N} w_i \cdot (y_i - \bar{y}) \cdot (\hat{y}_i - \bar{\hat{y}}) }{ \sqrt{ \sum_{i=1}^{N} w_i \cdot (y_i - \bar{y})^2 } \cdot \sqrt{ \sum_{i=1}^{N} w_i \cdot (\hat{y}_i - \bar{\hat{y}})^2 } }
    \label{eq:weighted-acc}
\end{equation}

where:
\begin{itemize}
    \item \( N \) is the total number of grid points,
    \item \( y_i \) is the observed value at grid point \( i \),
    \item \( \hat{y}_i \) is the predicted value at grid point \( i \),
    \item \( \bar{y} \) is the mean of the observed values,
    \item \( \bar{\hat{y}} \) is the mean of the predicted values,
    \item \( w_i \) is the latitude weight at grid point \( i \), calculated as:
    \[
    w_i = \cos(\phi_i)
    \]
    where \( \phi_i \) is the latitude at grid point \( i \).
\end{itemize}

High-resolution precipitation diagnosis was evaluated using both quantitative and qualitative approaches. Quantitative metrics included the Threat Score (TS), Probability of Detection (POD), and False Alarm Ratio (FAR). For qualitative assessment, we randomly selected case studies to examine the spatial distribution of precipitation. This comprehensive evaluation framework ensures a rigorous analysis of the model's effectiveness.

The calculations of TS, POD, and FAR are based on a contingency table, which categorizes forecast results into four categories. The TS, also known as the Critical Success Index (CSI), measures the fraction of correctly predicted events relative to the total number of events predicted or observed (the higher, the better), defined as \ref{eq:ts}. The POD measures the fraction of observed events that were correctly forecasted(the higher, the better), defined as \ref{eq:pod}. The FAR measures the fraction of forecasted events that did not occur(the lower, the better), defined as \ref{eq:far}.

\begin{table}[H]
    \centering
    \caption{Contingency Table for Precipitation Forecasting}
    \begin{tabular}{lcc}
        \toprule
        & \textbf{Observed} & \textbf{Not Observed} \\
        \midrule
        \textbf{Forecasted} & Hits & False Alarms \\
        \textbf{Not Forecasted} & Misses & True Negatives \\
        \bottomrule
    \end{tabular}
\end{table}

\begin{equation}
    TS=\frac{Hits}{Hits+False Alarms+Misses}
    \label{eq:ts}
\end{equation}
\begin{equation}
    POD=\frac{Hits}{Hits+Misses}
    \label{eq:pod}
\end{equation}
\begin{equation}
    FAR=\frac{False Alarms}{Hits+False Alarms}
    \label{eq:far}
\end{equation}

\section{Results}

\subsection{Regional Weather Forecast}

Latitude weighted RMSE and ACC are applied over China to assess the performance of both our CNCast model and Pangu. The reference climatology for each variable in this study was obtained by calculating the hourly mean from 1979 to 2019. These metrics are calculated for all surface variables and pressure level variables at 500 hPa, based on the test dataset from July 2021.

\subsubsection{Surface Weather Variables}

\begin{figure}[H]
    \centering
    \includegraphics[width=0.98\textwidth]{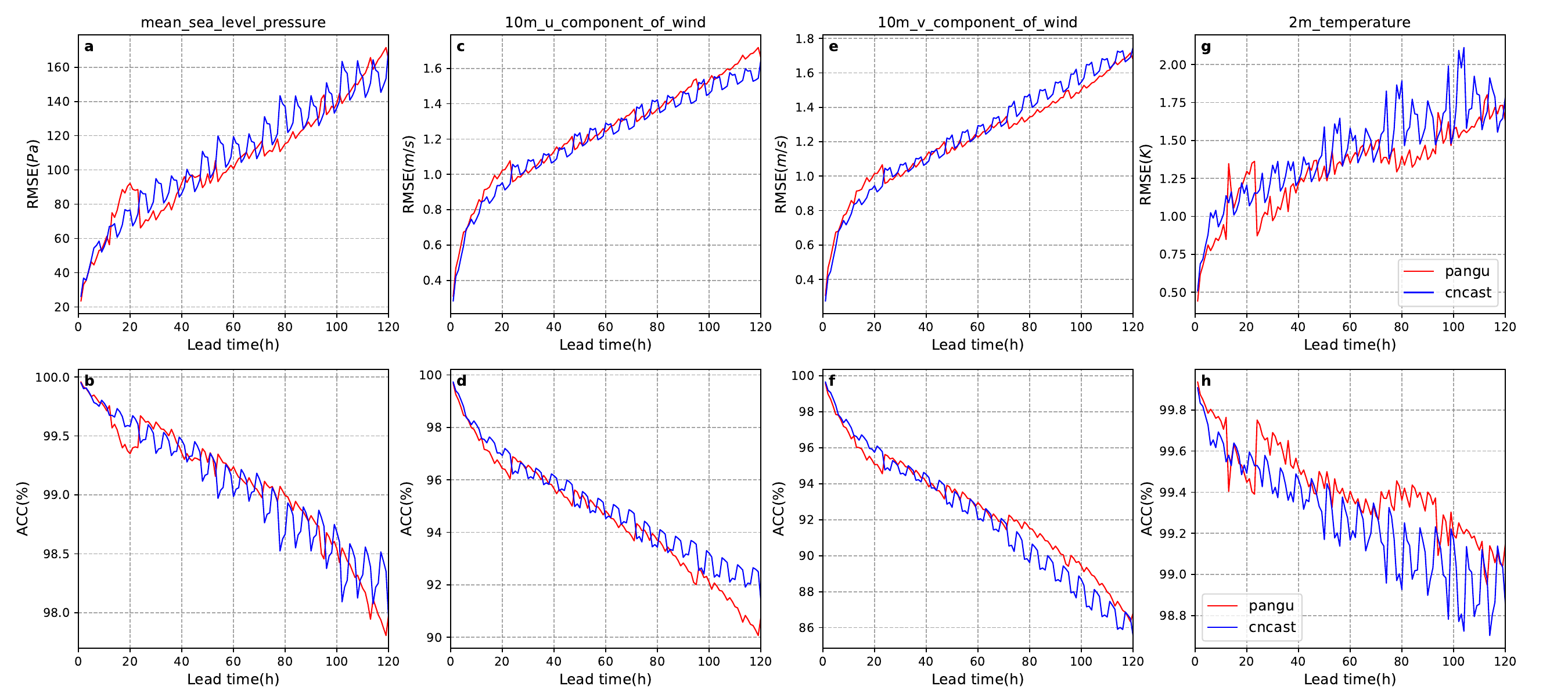} 
    \caption{The comparison of forecast accuracy in terms of latitude-weighted RMSE and ACC of 4 surface variables.}
    \label{fig:surface_rmse}
\end{figure}

Figure \ref{fig:surface_rmse} provides a detailed comparison of latitude-weighted RMSE and ACC between Pangu and CNCast for four surface variables: the u- and v-components of 10m wind, 2m temperature (2mt), and mean sea level pressure (mslp). The results reveal that CNCast outperforms in forecasting the 10m u-component and v-component of wind, as indicated by its lower RMSE and generally higher ACC values compared to Pangu across most lead times. This suggests that CNCast achieves higher accuracy in capturing the dynamics of near-surface wind fields. Conversely, the RMSE values for 2mt and mslp are slightly higher for CNCast than for Pangu, indicating that Pangu performs better in predicting these variables. 

In terms of ACC, CNCast demonstrates stronger anomaly correlation for the u- and v-components of wind at various lead times, further emphasizing its enhanced capability in wind-related forecasts. For 2mt and mslp, ACC values of CNCast displays smaller values than Pangu. These findings underscore CNCast's particular strength in wind prediction while maintaining solid performance across other surface variables. The figure highlights the nuanced strengths of each model, with CNCast emerging as a particularly reliable tool for wind forecasting applications. 

Additionally, both CNCast and Pangu exhibit an increase in RMSE (and a decrease in ACC) in a zig-zag pattern as lead time extends. This phenomenon is attributed to the use of a greedy strategy in the iterative forecasting process by both models. This strategy minimizes the number of iterations required for each lead time, resulting in the observed zig-zag behavior.

\begin{figure}[H]
    \centering
    \begin{subfigure}[b]{0.55\linewidth}
        \centering
        \includegraphics[width=\linewidth]{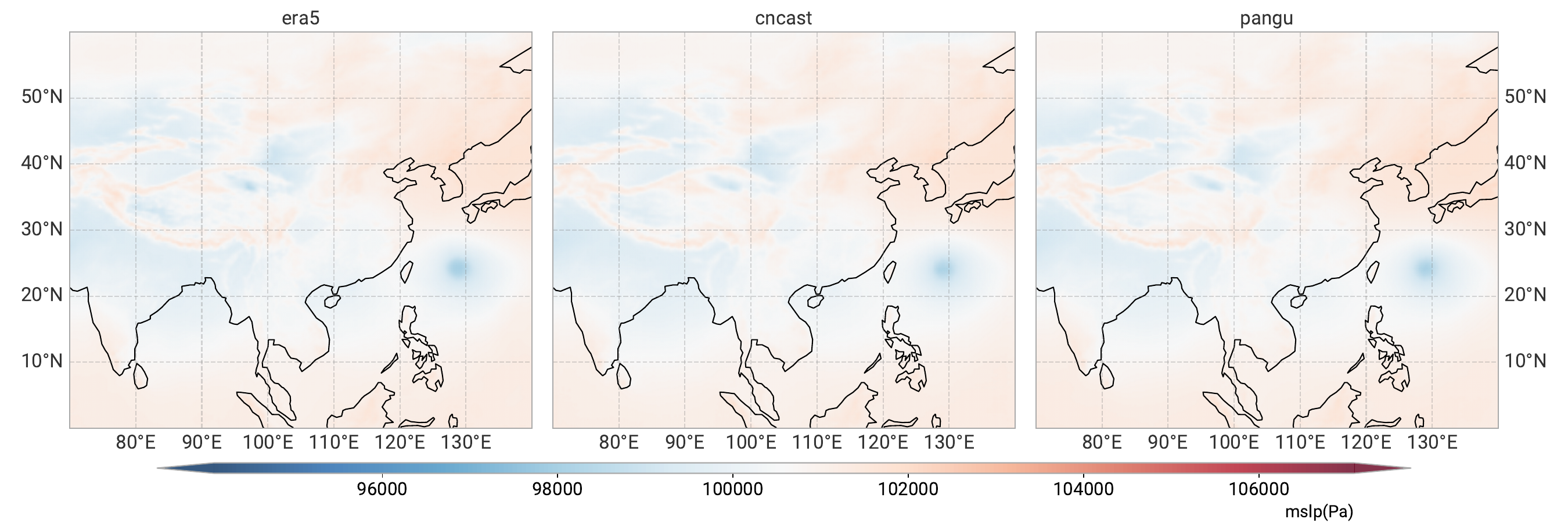}
        \label{fig:2mt}
    \end{subfigure}
    \begin{subfigure}[b]{0.38\linewidth}
        \centering
        \includegraphics[width=\linewidth]{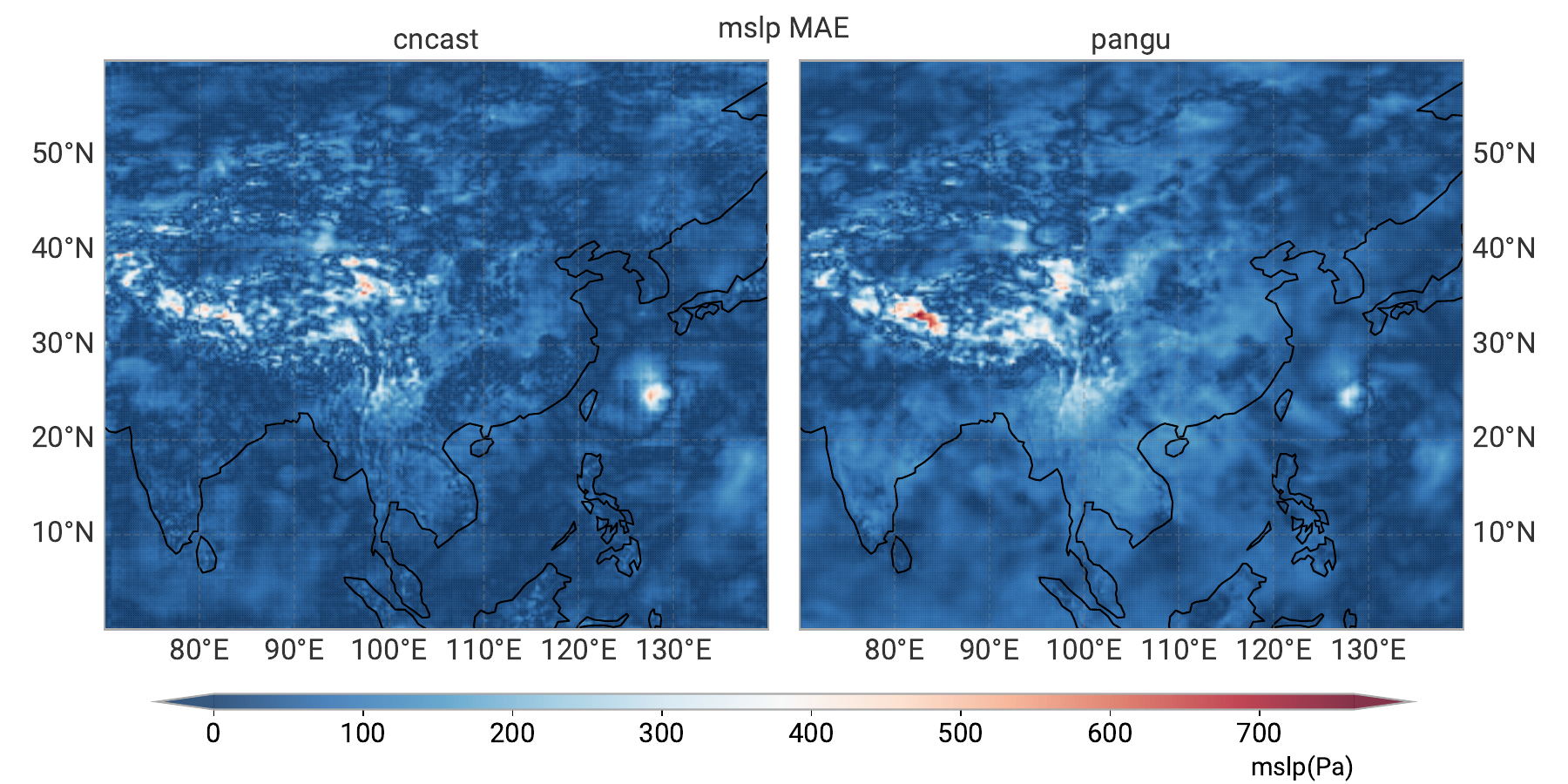}
        \label{fig:2mt_bias}
    \end{subfigure}
    \begin{subfigure}[b]{0.55\linewidth}
        \centering
        \includegraphics[width=\linewidth]{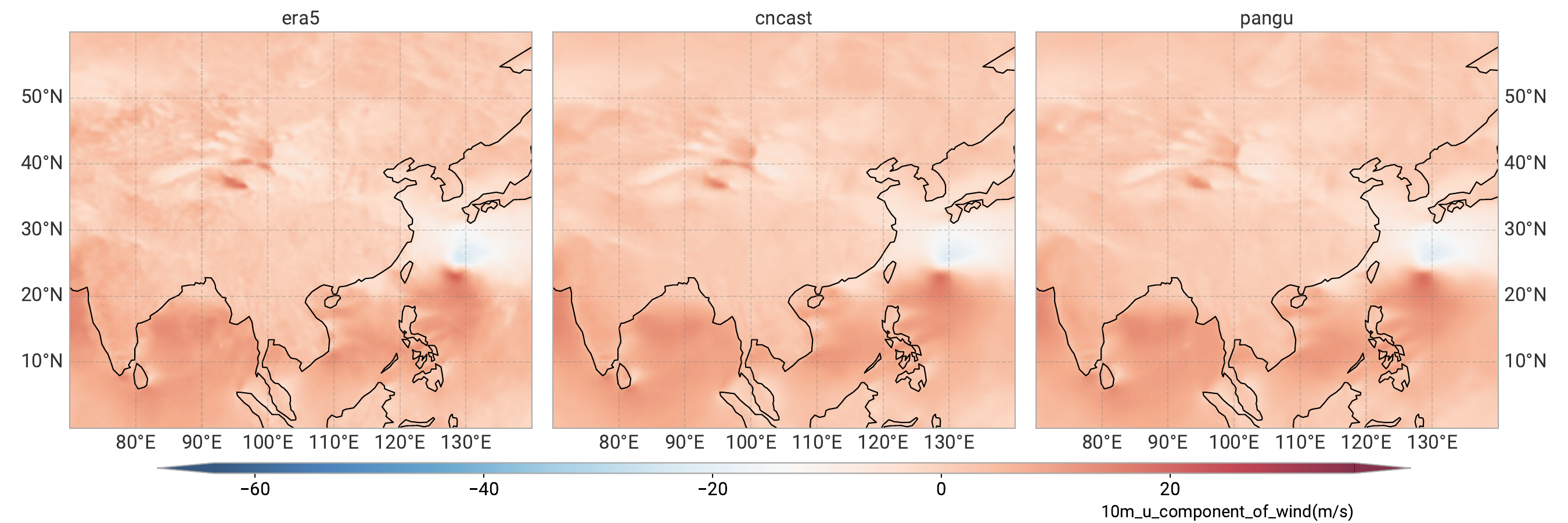}
        \label{fig:10u}
    \end{subfigure}
    \begin{subfigure}[b]{0.38\linewidth}
        \centering
        \includegraphics[width=\linewidth]{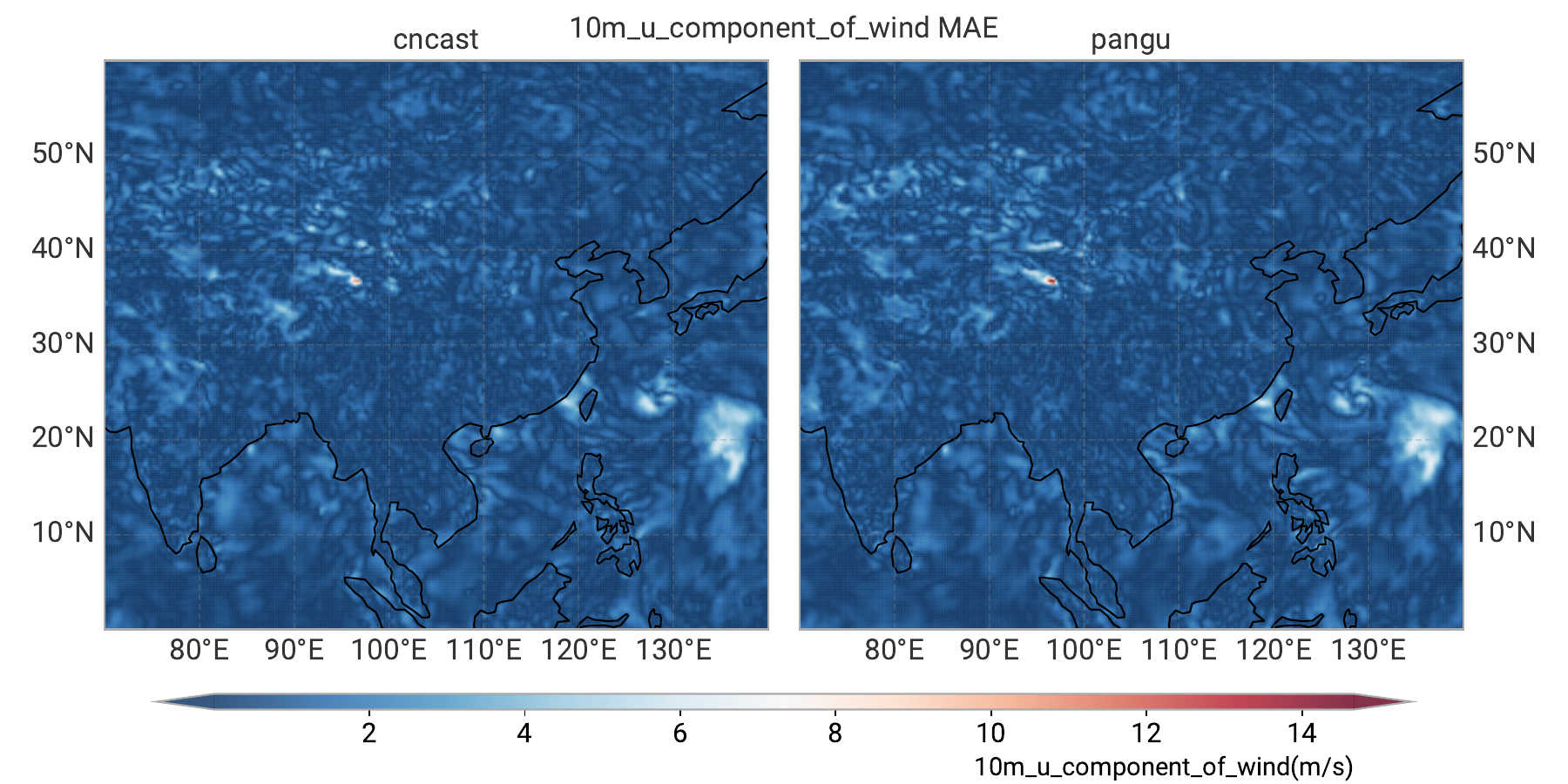}
        \label{fig:10u_bias}
    \end{subfigure}
    \caption{Visualization of 12 hour weather forecast and spatial MAE of mslp(top) and u-component of wind(bottom) produced by CNCast, Pangu and the ERA5 ground-truth. The initial time point (i.e. the forecast is performed on) is 2021072001 UTC.}
    \label{fig:surface_fcst}
\end{figure}

Figure \ref{fig:surface_fcst} illustrates the 12 hour forecasts for two near-surface variables, initialized at 2021072001. Both CNCast and Pangu demonstrate their ability to produce large-scale spatial distributions that closely align with the ERA5 reanalysis data. However, due to the inherent challenges and increased uncertainty associated with longer lead times, the forecast results exhibit some degree of blurring. Notably, the mslp and wind forecasts capture the presence of a typhoon off the eastern coast of China at 2021072301.

The mslp forecasts exhibit a mean absolute error(MAE) of approximately 1.5hPa, with the maximum MAE reaching 7hPa, primarily concentrated over the mountains around Tibetan Plateau. Additionally, CNCast exhibits a lower overall forecast MAE compared to CNCast in South Asia and the Indian Ocean, while its MAE is larger at typhoon center in the Pacific. 

In terms of the u-component of wind, the MAE is approximately 1m/s, with the maximum MAE reaching 14 m/s, particularly pronounced in typhoon-active regions. These findings highlight both the strengths and limitations of the models. CNCast tends to exhibit lower MAE in certain areas, especially in complex terrain and during typhoon-related events. Conversely, Pangu shows relatively poorer performance in these challenging scenarios. This suggests that CNCast demonstrates superior wind forecasting performance in mountainous regions and typhoon zones. 

\subsubsection{Pressure Level Variables}

\begin{figure}[H]
    \centering
    \includegraphics[width=0.98\textwidth]{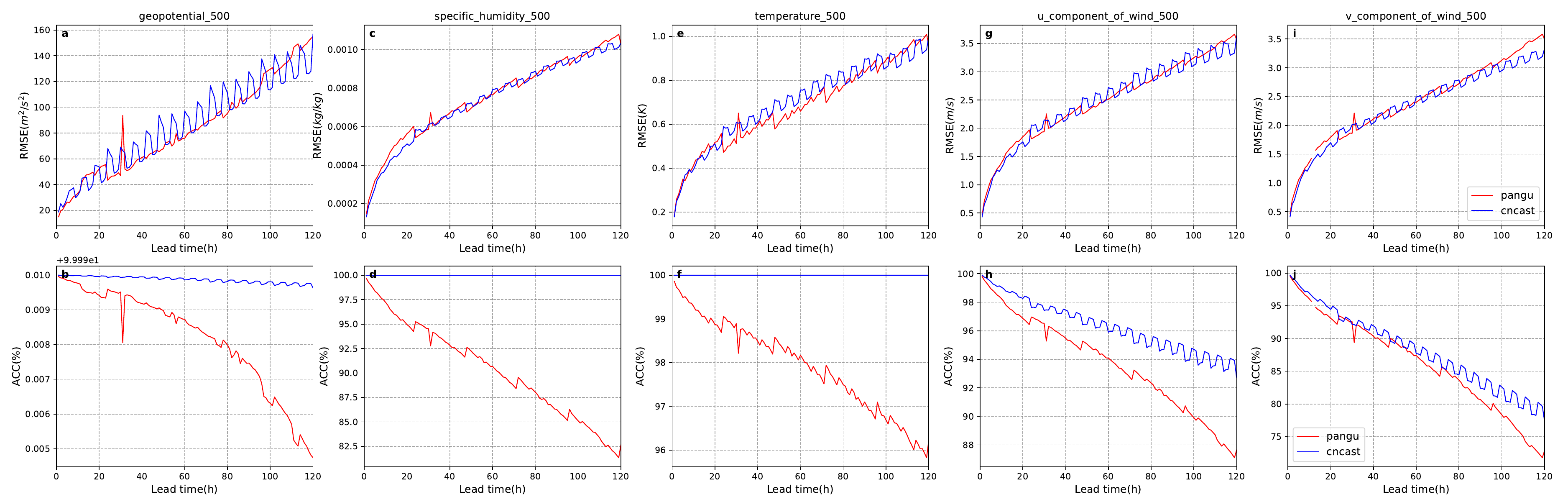} 
    \caption{The comparison of forecast accuracy in terms of latitude-weighted RMSE and ACC of pressure level variables at 500hPa.}
    \label{fig:upper_air_500hpa}
\end{figure}

Figure \ref{fig:upper_air_500hpa} presents the latitude-weighted RMSE of five pressure level variables at 500 hPa level: temperature, geopotential, specific humidity, and the u- and v-components of wind, comparing Pangu and CNCast. The analysis reveals consistent trends in RMSE and ACC across all variables as lead time increases. Specifically, RMSE values tend to rise, while ACC values decline, reflecting the inherent challenges of long-term forecasting for atmospheric variables. This pattern aligns with the expected degradation of predictive skill over extended lead times, a well-documented phenomenon in numerical weather prediction.

Additionally, similar to the metrics observed for surface variables, both RMSE and ACC exhibit a zigzag pattern as lead time increases. This behavior is attributed to the greedy strategy implemented in iterative forecasting, which influences the metrics' variability over different lead times.

For the variables including the u- and v-components of wind, temperature, and specific humidity, CNCast demonstrates superior performance, evidenced by generally lower RMSE values and higher ACC values compared to Pangu. This highlights CNCast's enhanced capability to accurately capture the dynamics and variability of these critical atmospheric parameters. However, when it comes to geopotential, CNCast shows higher RMSE and lower ACC across nearly all lead times compared to Pangu, indicating relatively poorer performance. This discrepancy may be attributed to the complex terrain of the Tibetan Plateau within the Chinese region, which presents significant variations in potential height. These variations pose greater modeling challenges compared to global potential heights, potentially impacting CNCast's ability to accurately predict geopotential in these areas.

Overall, in the China region, CNCast demonstrates superior performance in forecasting wind and specific humidity, while Pangu excels in predicting geopotential, near-surface temperature, and mean sea level pressure. This suggests that CNCast may hold a distinct advantage in applications such as wind power forecasting and moisture-related predictions, where accurate simulation of wind fields and humidity is critical. Conversely, Pangu's strengths in geopotential and surface variables underscore its robustness in capturing large-scale atmospheric dynamics and boundary layer processes. These complementary strengths highlight the potential for leveraging both models synergistically to address diverse forecasting needs across different meteorological applications. By integrating the unique capabilities of each model, it may be possible to enhance overall predictive performance and better meet the specific requirements of various weather-dependent sectors.

\begin{figure}[H]
    \centering
    \begin{subfigure}[b]{0.55\linewidth}
        \centering
        \includegraphics[width=\linewidth]{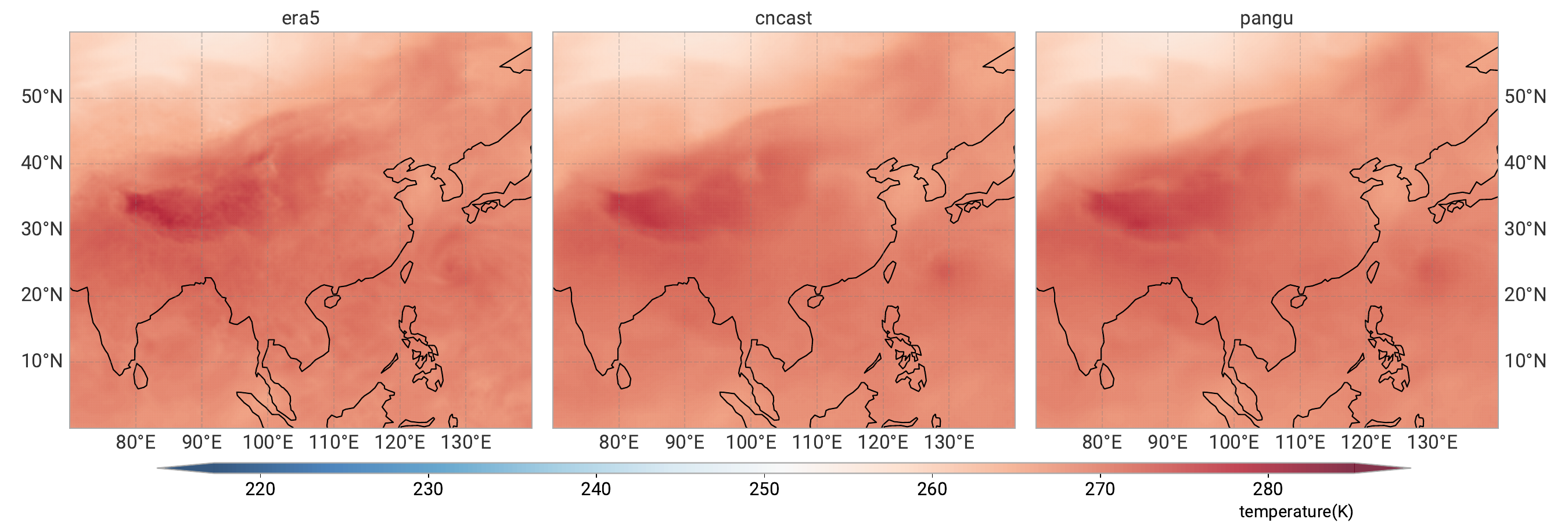}
        \label{fig:t500}
    \end{subfigure}
    \begin{subfigure}[b]{0.38\linewidth}
        \centering
        \includegraphics[width=\linewidth]{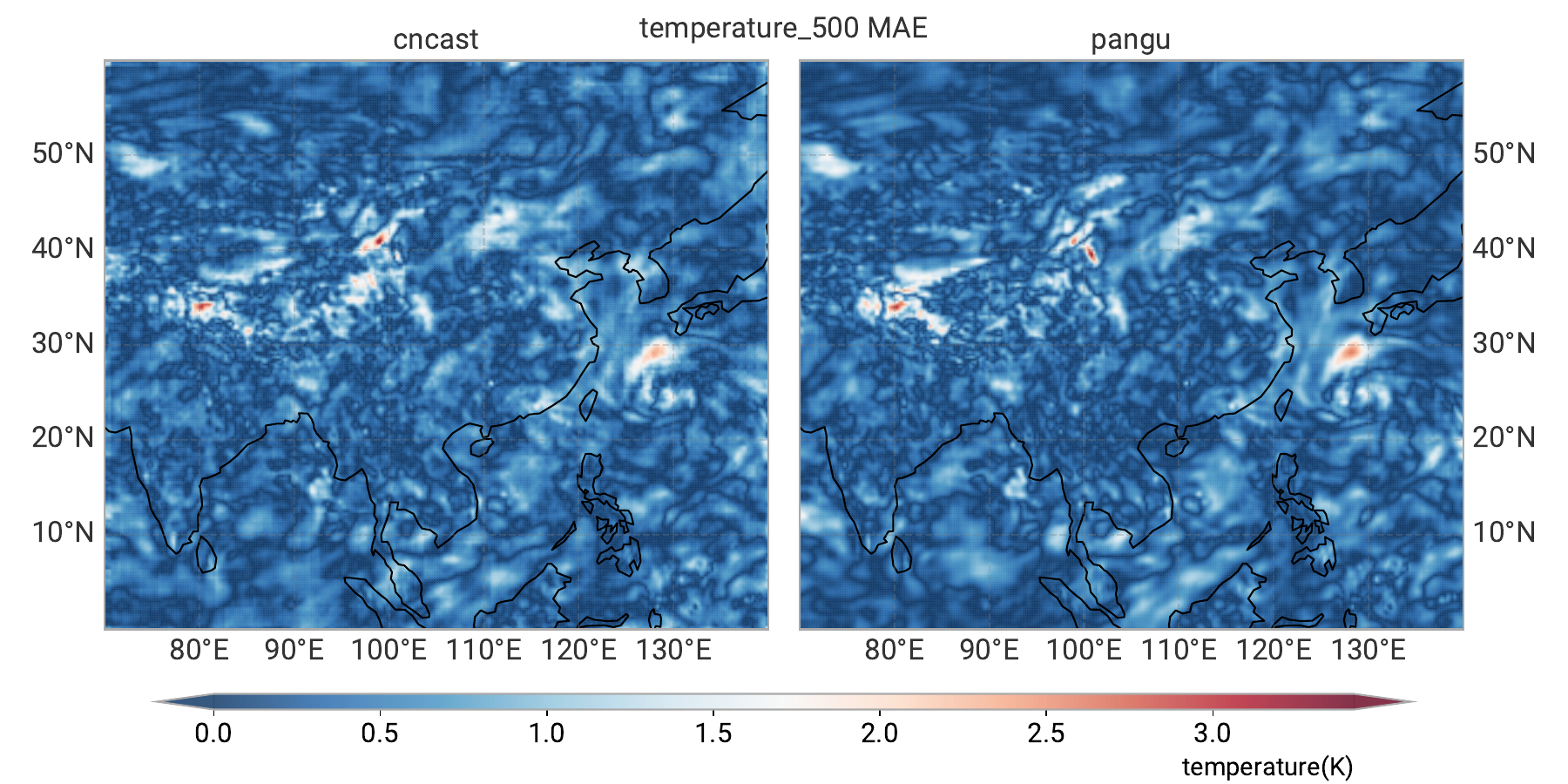}
        \label{fig:t_bias}
    \end{subfigure}
    \begin{subfigure}[b]{0.55\linewidth}
        \centering
        \includegraphics[width=\linewidth]{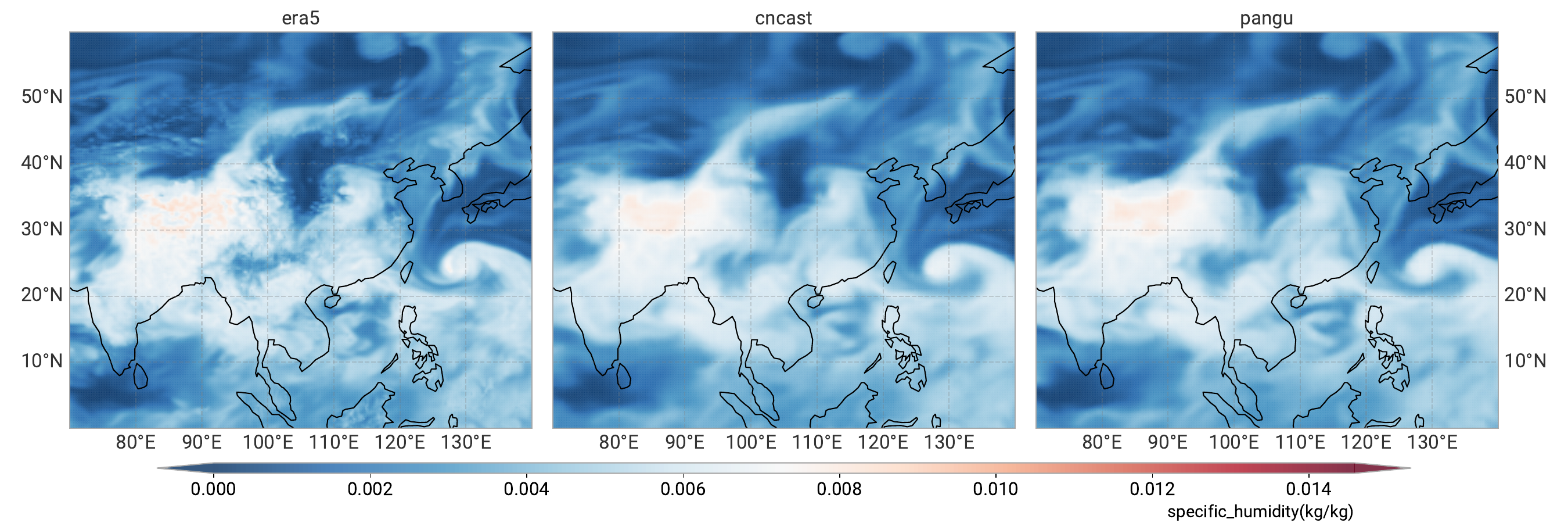}
        \label{fig:q500}
    \end{subfigure}
    \begin{subfigure}[b]{0.38\linewidth}
        \centering
        \includegraphics[width=\linewidth]{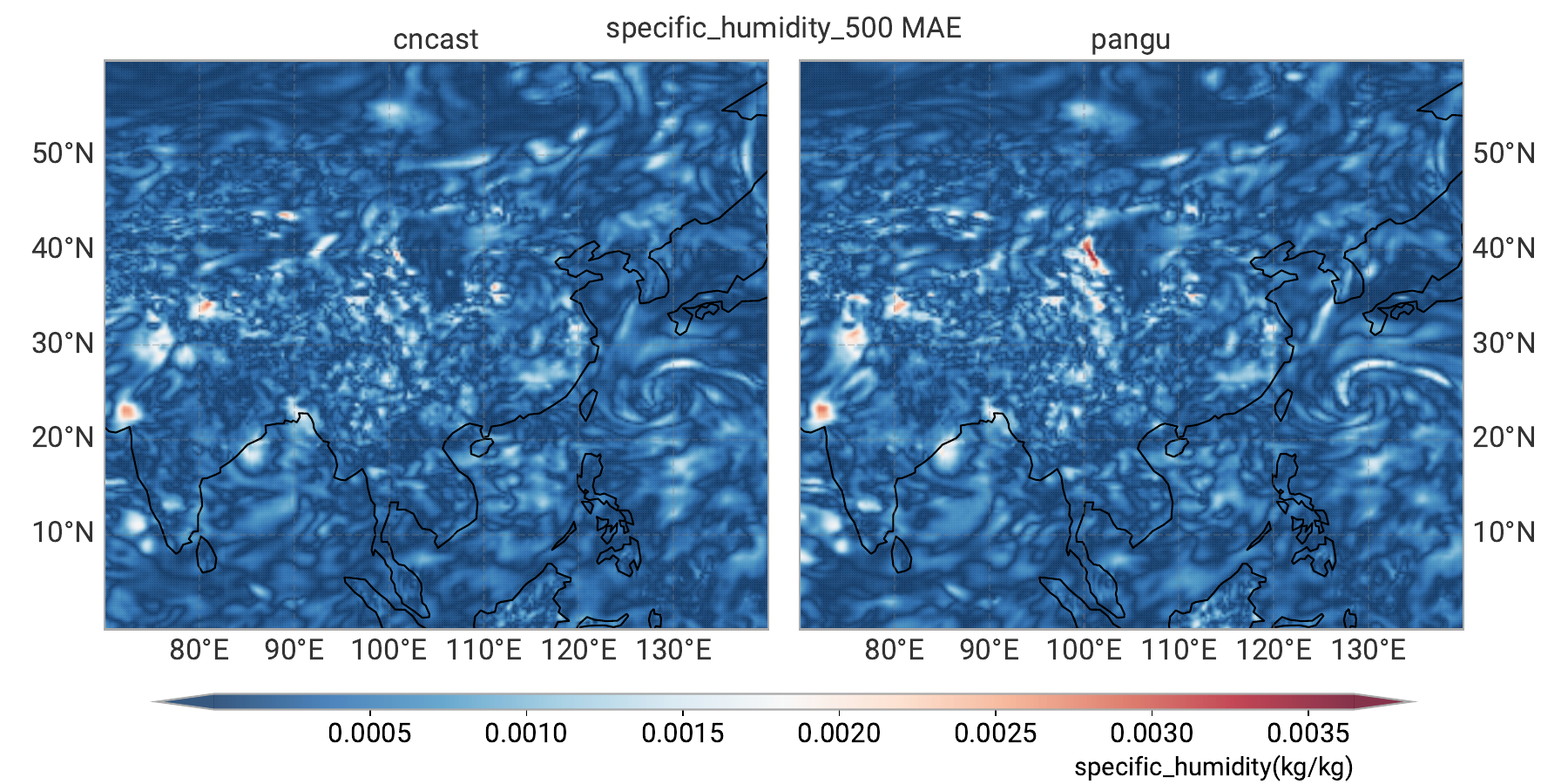}
        \label{fig:q_bias}
    \end{subfigure}
    \begin{subfigure}[b]{0.55\linewidth}
        \centering
        \includegraphics[width=\linewidth]{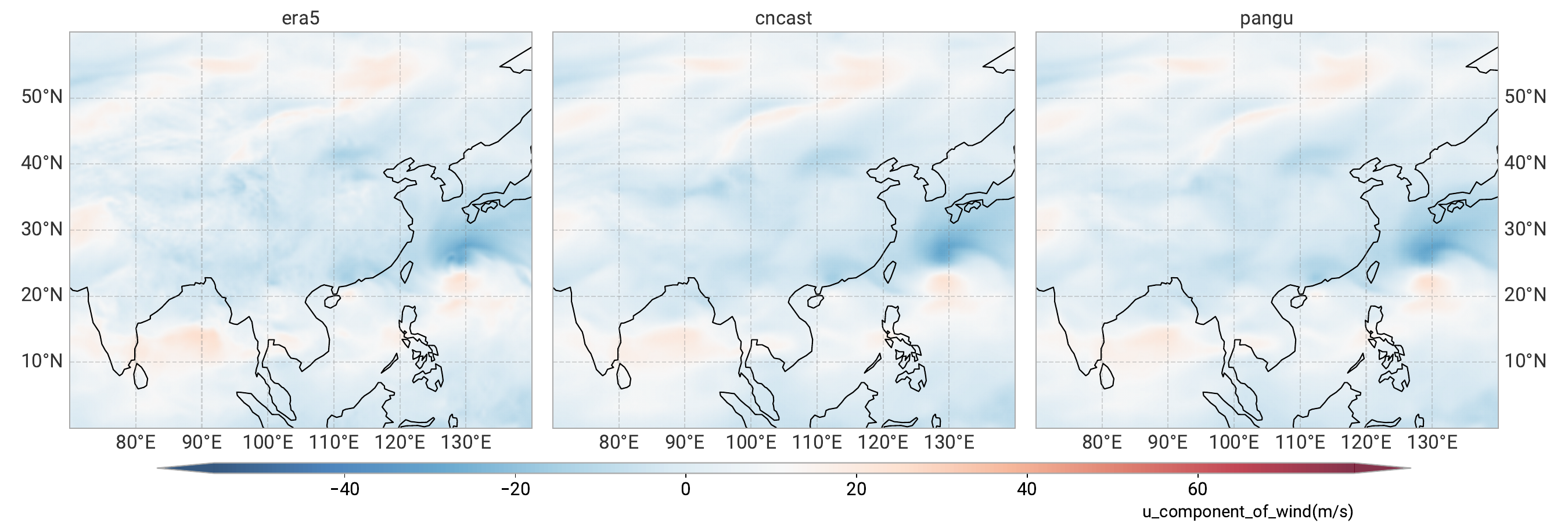}
        \label{fig:u500}
    \end{subfigure}
    \begin{subfigure}[b]{0.38\linewidth}
        \centering
        \includegraphics[width=\linewidth]{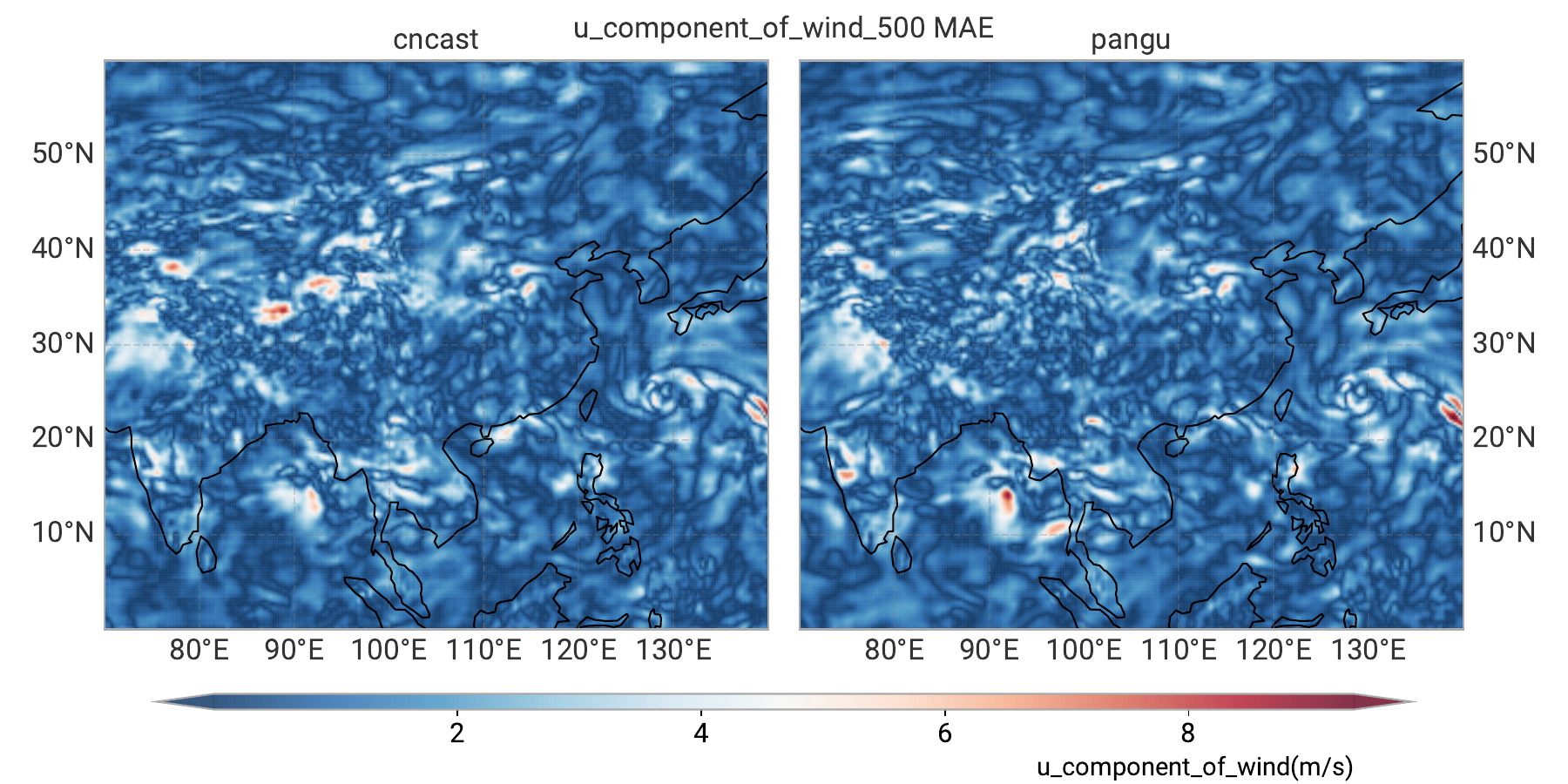}
        \label{fig:u_bias}
    \end{subfigure}
    \caption{Visualization of 12 hour weather forecast and spatial MAE of z500(top), q500(middle) and u500(bottom) variables produced by CNCast, Pangu and the ERA5 ground-truth. The initial time point (i.e. the forecast is performed on) is 2021072001 UTC.}
    \label{fig:pressure level_fcst}
\end{figure}

Figure \ref{fig:pressure level_fcst} illustrates the spatial distribution of 12 hour forecasts and MAE for three pressure level variables at 500hPa —temperature(top), specific humidity(middle), and the u-component of wind(bottom)—initialized at 2021072001. Similar to the near-surface variables, both CNCast and Pangu generate large-scale spatial distributions that closely align with the ERA5 reanalysis data. However, as with the near-surface forecasts, these results exhibit increased blurring and reduced texture clarity, indicative of the growing uncertainty and inherent challenges associated with longer lead times. This degradation in detail is a common characteristic of extended-range forecasts, as small-scale features become more difficult to resolve accurately over time.

Despite these challenges, the models maintain a reasonable representation of the overall atmospheric patterns, demonstrating their utility in capturing large-scale dynamics. This ability underscores the model's effectiveness in providing valuable insights into large-scale weather systems.

\subsection{High Resolution Precipitation Diagnosis}

High-resolution precipitation was assessed using data from 2021. After training the DiT model, latent representations of ERA5 variables, such as temperature, pressure, specific humidity, and wind, along with the latent representation of CMPAS at the previous time step $t-1$, were input into the model to estimate the latent representation of CMPAS precipitation at the current time step. This latent output was then processed by the CMPAS decoder to generate the precipitation field.

To produce more realistic precipitation fields, we set the number of diffusion steps to 250 during inference. The diffusion model is capable of generating diverse outputs by initializing with different noise patterns, which enables ensemble-based improvements in diagnostic accuracy. To leverage this capability, we generated three distinct ensemble members with the diffusion model and applied the EnMax ensemble method\cite{enmax} to integrate them and to obtain the final results.

\subsubsection{Metrics}

\begin{figure}[H]
    \centering
    \begin{subfigure}[b]{0.3\linewidth}
        \includegraphics[width=\linewidth]{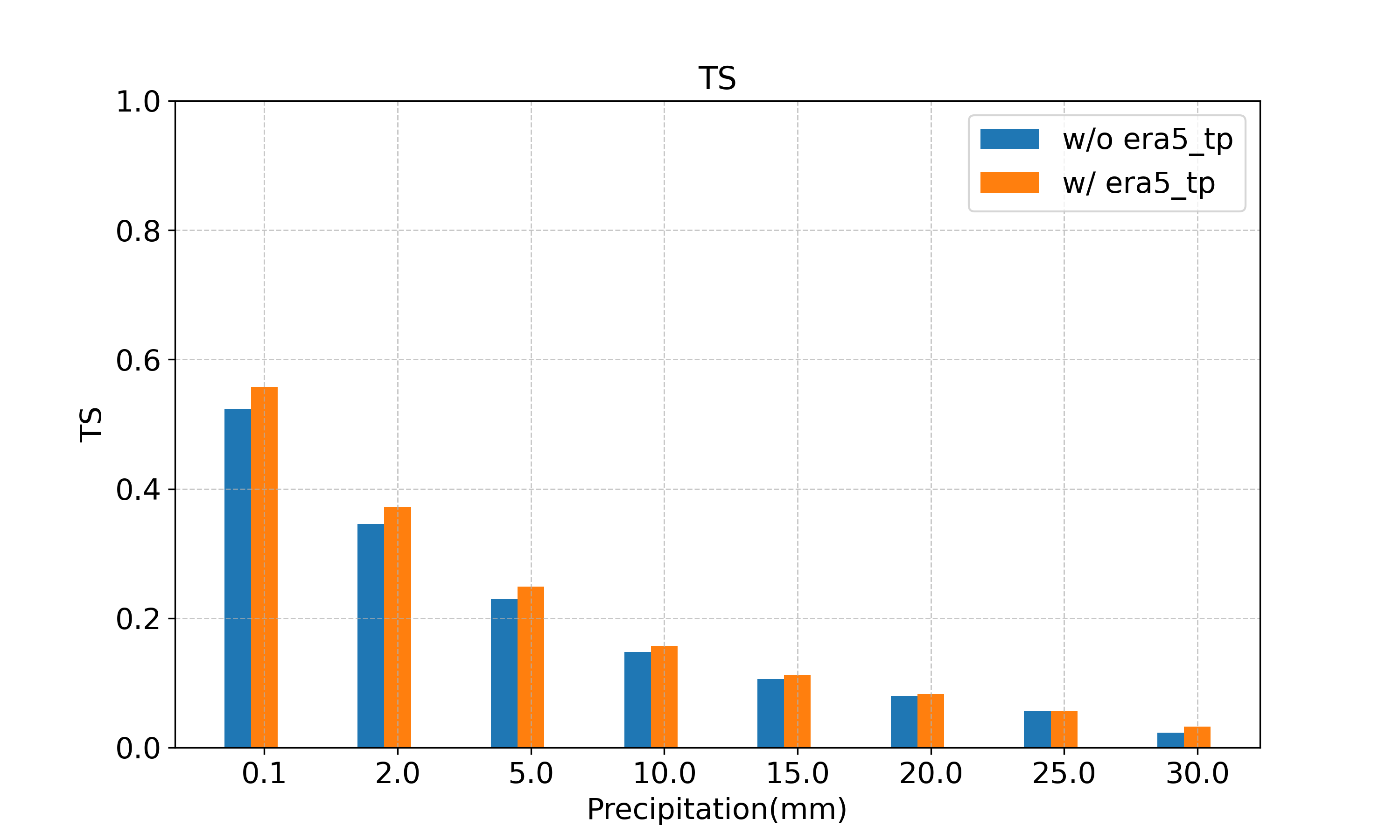}
        \label{fig:ts}
    \end{subfigure}
    \begin{subfigure}[b]{0.3\linewidth}
        \includegraphics[width=\linewidth]{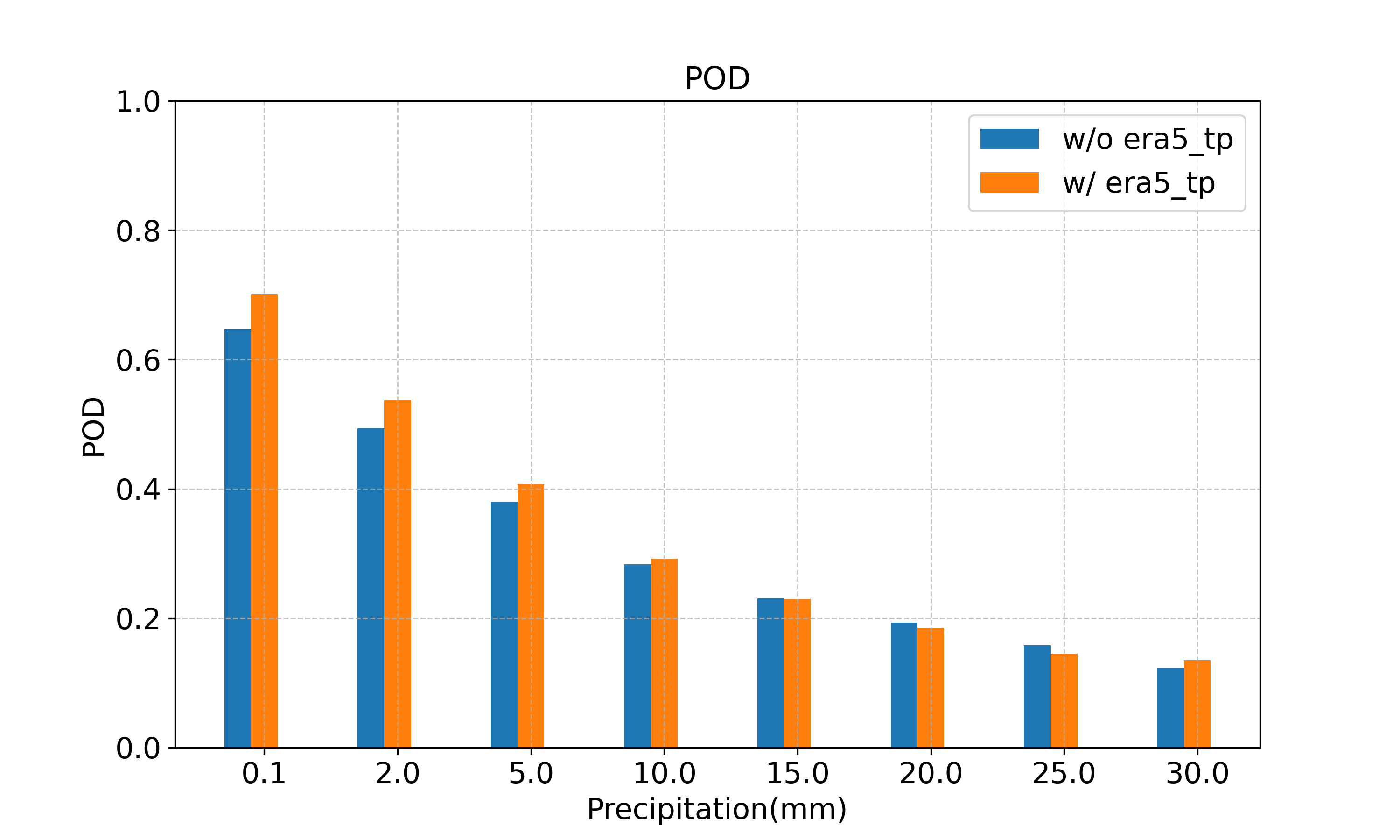}
        \label{fig:pod}
    \end{subfigure}
    \begin{subfigure}[b]{0.3\linewidth}
        \includegraphics[width=\linewidth]{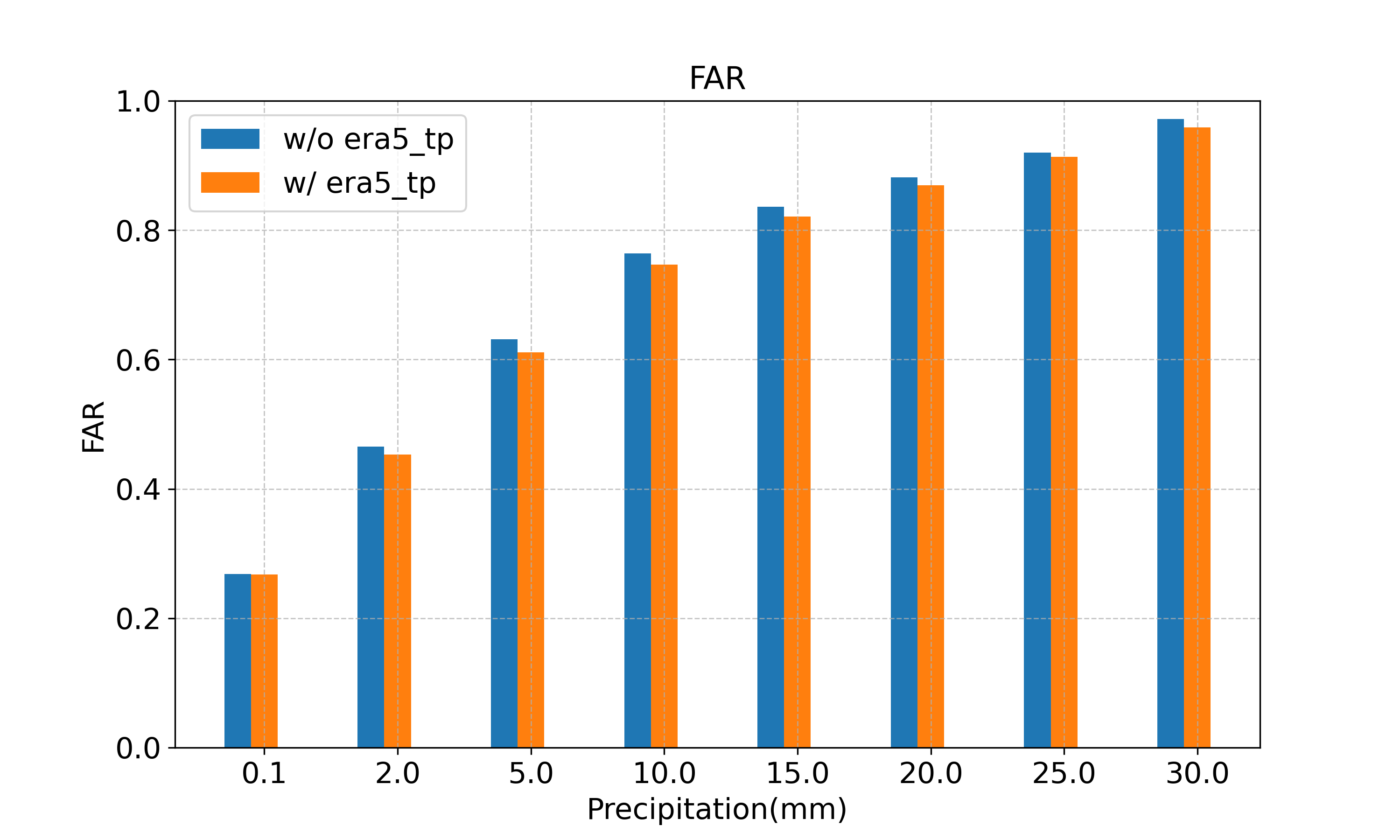}
        \label{fig:far}
    \end{subfigure}
    \caption{TS, POD and FAR of estimated precipitation across 2021.}
    \label{fig:precip metrics}
\end{figure}

Figure \ref{fig:precip metrics} presents the TS, POD, and FAR for hourly accumulated precipitation at various thresholds, calculated using CMPAS hourly precipitation as the ground truth based on 2021 data. The results show that as the precipitation threshold increases, both TS and POD scores tend to decline, while the FAR increases. For distinguishing between rainy and non-rainy conditions, the TS score reaches 0.55, with a POD of 0.7. In the case of heavy rainfall ($\geq$10 mm), the TS score reaches 0.16, the POD exceeds 0.28, and the FAR is approximately 0.7. These metrics highlight the challenges in forecasting more intense precipitation events, where accuracy tends to decrease.

Furthermore, incorporating ERA5 precipitation features significantly enhances the TS score, demonstrating a notable improvement in the model’s ability to diagnose precipitation compared to models that do not include ERA5 precipitation data. This integration underscores the value of using comprehensive datasets to improve predictive accuracy in complex meteorological scenarios.

\subsubsection{Case Study}

\begin{figure}[H]
    \centering
    \begin{subfigure}[b]{0.45\linewidth}
        \includegraphics[width=\linewidth]{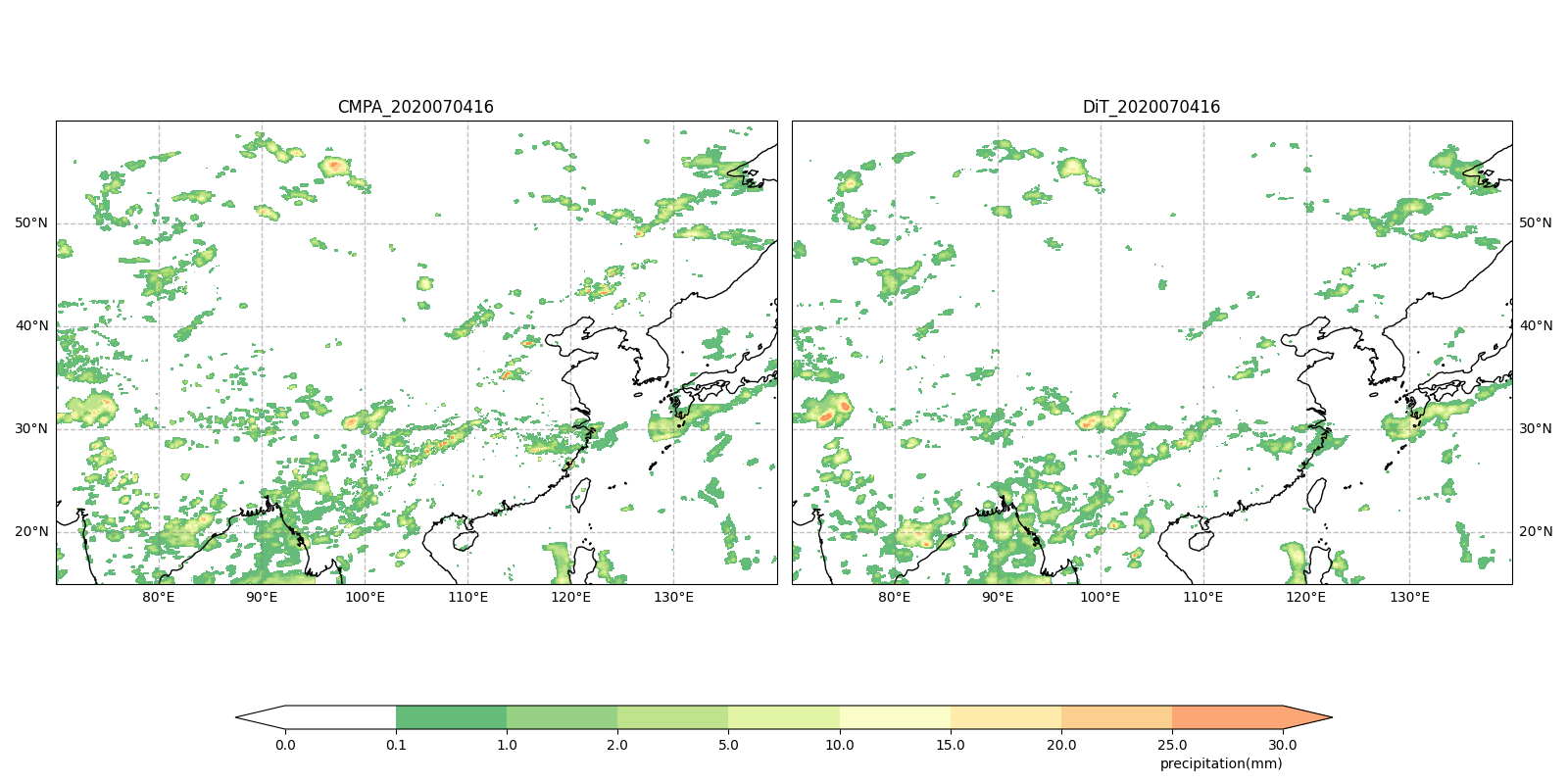}
        \label{fig:prcp_2020070416}
    \end{subfigure}
    \begin{subfigure}[b]{0.45\linewidth}
        \includegraphics[width=\linewidth]{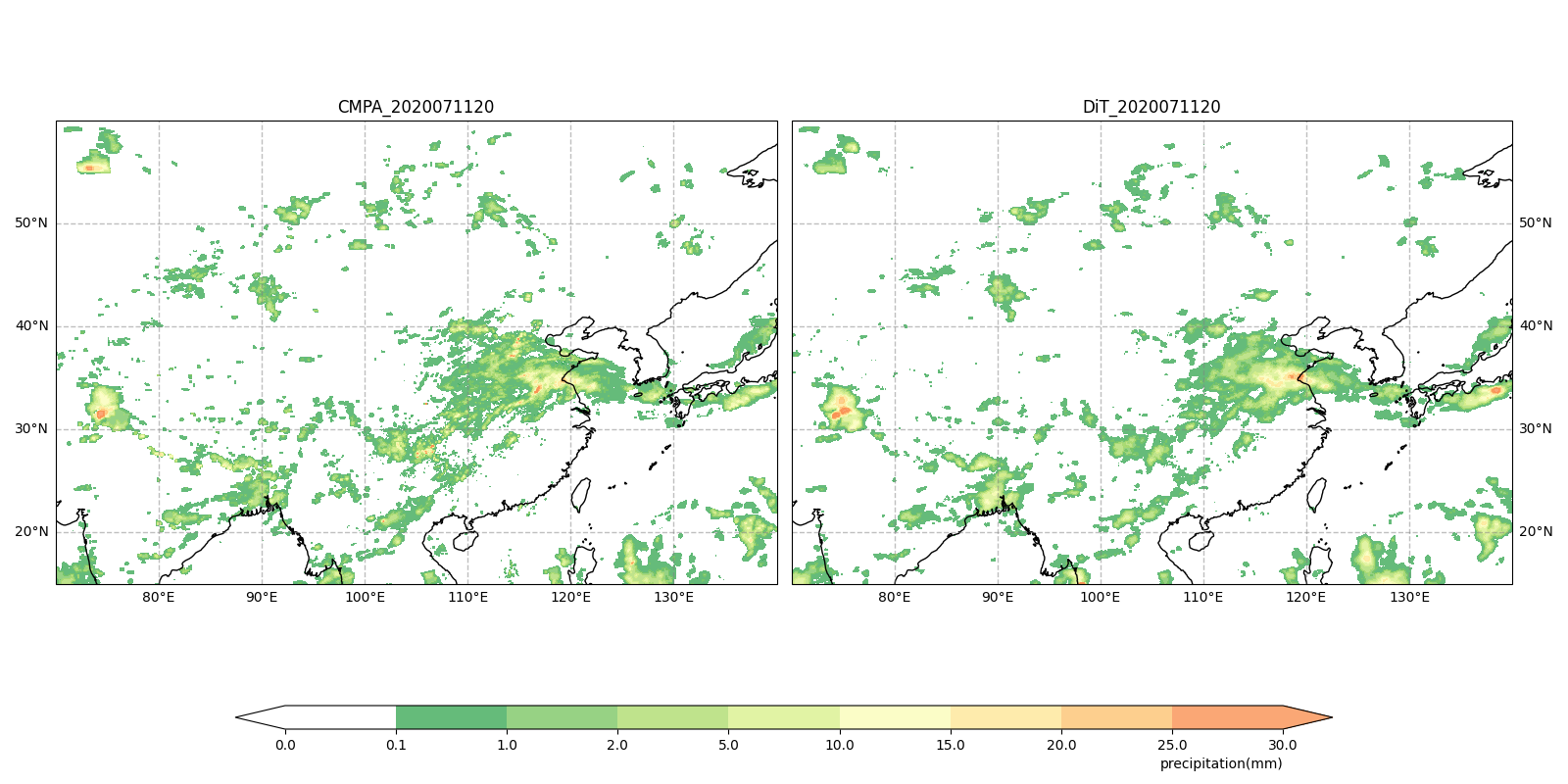}
        \label{fig:prcp_2020071120}
    \end{subfigure}
    \caption{Visualization of the estimated and CMPAS ground-truth high resolution precipitation at 2020070416(left) and 2020071120(right), respectively.}
    \label{fig:precip distributions}
\end{figure}

Figure \ref{fig:precip distributions} compares high-resolution precipitation diagnosed by the DiT model with CMPAS precipitation. The left panel corresponds to 16 UTC on July 4, 2020, while the right panel represents 20 UTC on July 11, 2020. The results indicate that the DiT model effectively captures the spatial distribution of precipitation, particularly for large-scale precipitation systems. It provides reasonable estimates of the centers of intense, large-scale precipitation. However, the model tends to underestimate scattered small-scale, high-intensity precipitation. Despite utilizing a diffusion-based generative model, the forecasted precipitation texture lacks sharpness, likely due to the limited availability of high-resolution information.

These findings suggest that while the DiT model is proficient in modeling widespread precipitation patterns, there is room for improvement in resolving finer-scale features and enhancing the clarity of precipitation textures. Further refinement of input data resolution and model parameters could potentially address these limitations.

\section{Conclusions and Future Remarks}

This paper introduces CNCast, a regional forecasting framework specifically designed for China. CNCast harnesses an enhanced Swin Transformer 3D model, incorporating boundary conditions to provide hourly forecasts up to five days ahead. It effectively predicts key meteorological variables such as temperature, pressure, humidity, and wind, and has demonstrated generally superior forecasting accuracy over Pangu, particularly excelling in wind predictions.

In addition, a high-resolution precipitation diagnostic model was developed for China using CMPAS data as target, based on the latent diffusion model DiT. This diagnostic framework can be integrated with any forecasting model capable of covering China region, delivering high-resolution precipitation forecasts.

Despite these advancements, the current model encounters challenges in accurately forecasting temperature and geopotential, with forecast clarity diminishing as lead time increases. As generative models like diffusion, denoising score matching, and flow matching continue to advance in image and video generation, future work may explore these techniques to develop models with enhanced texture clarity and ensemble forecasting capabilities. This could lead to more precise and practical weather predictions, addressing current limitations and expanding the utility of regional forecasting models.

\acknowledgments
The authors would like to thank ECMWF for providing free access to the ERA5 reanalysis data and also express their gratitude to Google Cloud for facilitating the storage of multi-year data in the user-friendly Zarr format within a free storage bucket.

\section*{Data Availability Statement}

The ERA5 reanalysis data used in this study is publicly available at \url{https://cloud.google.com/storage/docs/public-datasets/era5?hl=zh-cn}. The fused precipitation data CMPAS used in this study could be bought from National Meteorological Information Center. Analyses are performed and result figures are created using Python (v3.10.1). 

%
%

\bibliography{regional_weather_forecast_model}

%
%
%
%
%

\medskip
\end{document}